\documentclass[conference]{IEEEtran}
\IEEEoverridecommandlockouts

\usepackage{cite}
\usepackage{amsmath,amssymb,amsfonts}
\usepackage{algorithmic}
\usepackage{graphicx}
\usepackage{textcomp}
\usepackage{xcolor}
\usepackage{hyperref}

\usepackage[caption=false,font=footnotesize]{subfig}
\usepackage{placeins}

\def\BibTeX{{\rm B\kern-.05em{\sc i\kern-.025em b}\kern-.08em
    T\kern-.1667em\lower.7ex\hbox{E}\kern-.125emX}}

\begin{document}

\title{
AsyncMixVLA: Runtime Asynchronous Robot Policy Switching for Adaptation to Dynamic Environments 
}

\author{\IEEEauthorblockN{\vspace{-1.5em}}}
\author{
Tamoghno Das,
Keyon Jazayeri,
Yerin Cho,
Nazmul Takbir,
Arshia Aravinthan,
Nikil D. Dutt
}

\maketitle
\vspace{-5mm}

\begin{abstract}
Vision-language-action (VLA) models translate visual observations and language instructions into robot actions, enabling general-purpose robotic control. However, environmental perturbations, such as objects moving during execution, can undermine their reliability. Different policies exhibit complementary failures under mid-rollout perturbations, motivating selective switching between policies for recovery. Such switching can stall the robot during inference of the newly invoked policy or, if performed asynchronously, act on stale observations.

We present \textbf{AsyncMixVLA}, an adaptive inference framework for proactive, asynchronous switching across heterogeneous VLA policies. A learned trigger balances recovery potential against intervention harm to selectively escalate to a higher-compute policy. AsyncMixVLA overlaps its inference with committed outgoing actions and aligns its context to the anticipated takeover state. On LIBERO-Long benchmark augmented with mid-rollout perturbations, AsyncMixVLA improves perturbed task success from \textbf{69.5\% to 77.1\%} over the lightweight policy. It achieves \textbf{84.9\% overall success}, comparable to 84.5\% for synchronous switching, while reducing median exposed handoff latency from \textbf{226.5\,ms to 0.1\,ms}. Without retraining or recalibration, it achieves \textbf{75.1\% success} across three unseen perturbation families, compared with 71.3\% for the stronger constituent policy in aggregate.

\end{abstract}







%
%


%


%
%

\section{Introduction}
\label{sec:introduction}

Vision-language-action (VLA) models map visual observations and language
instructions to robot actions, enabling a single policy to perform diverse
manipulation tasks~\cite{zitkovich2023rt2,kim2024openvla,black2025pi0}.
However, strong nominal performance does not guarantee reliable execution under
deployment shifts: VLA policies remain sensitive to changes in adjoining environment~\cite{fei2025liberoplus}.
Mid-task perturbations are particularly challenging because they can
invalidate the state on which ongoing actions were generated. Robust deployment
therefore requires both effective recovery and timely delivery of corrective
actions.

\begin{figure}
    \centering
    \includegraphics[width=0.9\linewidth]{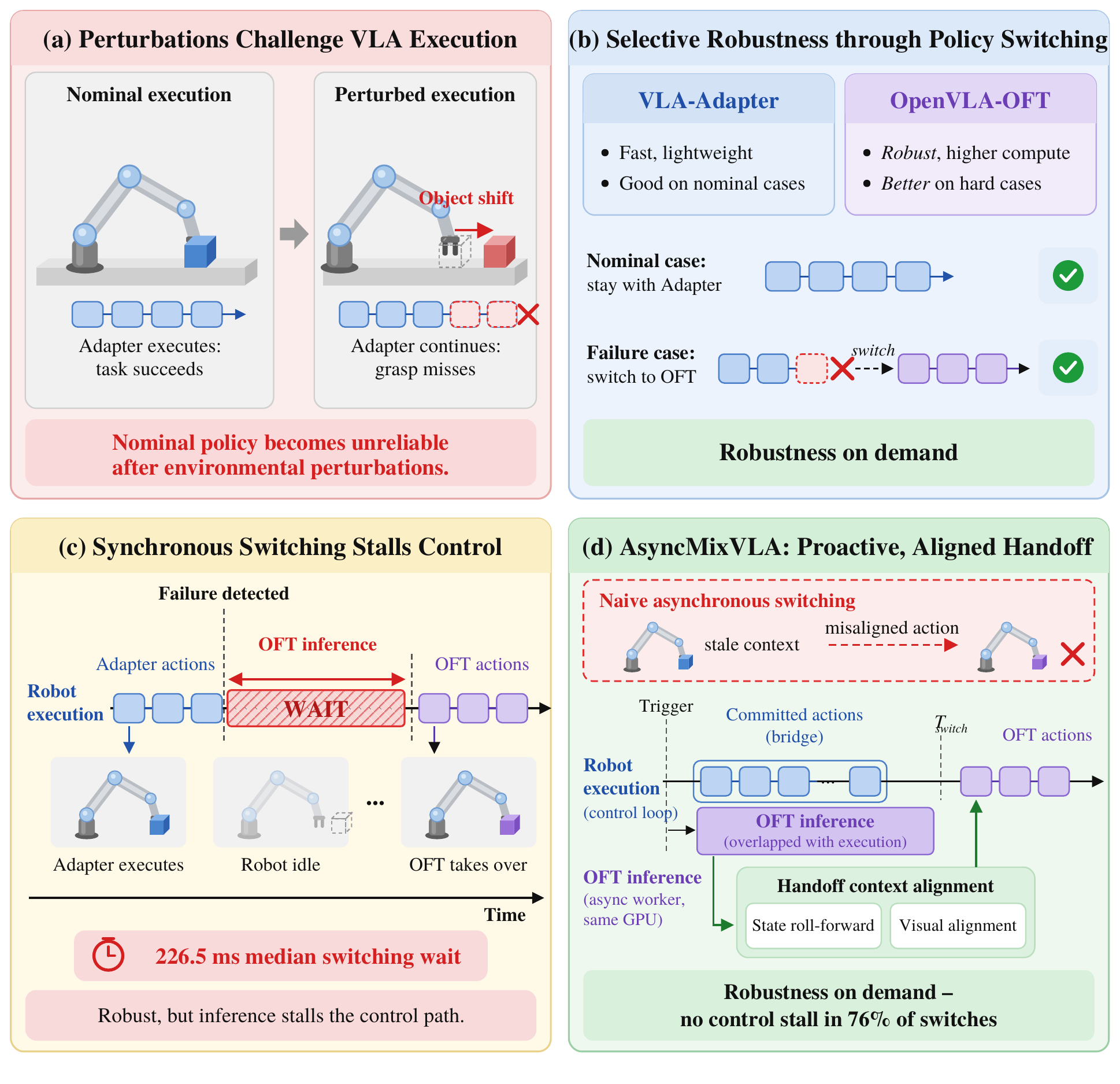}
    \caption{
    \textbf{Motivation for AsyncMixVLA.}
    \textbf{(a)}~Mid-rollout perturbations can make the lightweight
    VLA-Adapter unreliable.
    \textbf{(b)}~Heterogeneous VLA policies exhibit complementary failures,
    motivating selective escalation.
    \textbf{(c)}~Synchronous switching exposes incoming-policy inference
    latency to robot control.
    \textbf{(d)}~AsyncMixVLA proactively launches OpenVLA-OFT, overlaps
    inference with committed VLA-Adapter actions, and aligns its context to
    the anticipated takeover state.
    }
    \label{fig:motivation_fig}
\end{figure}

As shown in Fig.~\ref{fig:motivation_fig}, heterogeneous VLA policies can
exhibit complementary failure modes, motivating selective switching rather
than continuous use of a single policy, or switching to a compressed version of the same policy. Table~\ref{tab:model_complementarity}
confirms this empirically: VLA-Adapter~\cite{wang2025vlaadapter} recovers
substantially more OpenVLA-OFT~\cite{kim2025oft} failures than the
same-family OFT-Exit16 (Early exit \cite{song2025ceedvla} applied to OpenVLA-OFT) variant. We therefore use VLA-Adapter as the default
policy and selectively invoke OpenVLA-OFT for recovery.

\begin{table}[!t]
\centering
\caption{
\textbf{Policy complementarity on 500 matched perturbed LIBERO-Long rollouts.}
Failure coverage measures success on the 131 episodes failed by
OpenVLA-OFT; Best-of-Pair SR denotes success when either policy succeeds.
}
\label{tab:model_complementarity}
\setlength{\tabcolsep}{4.2pt}
\renewcommand{\arraystretch}{1.10}
\resizebox{\columnwidth}{!}{%
\begin{tabular}{lcc}
\hline
\textbf{Companion to OpenVLA-OFT}
&
\textbf{Failure Coverage}
&
\textbf{Best-of-Pair SR}
\\
\hline
\textbf{VLA-Adapter}
& \textbf{54/131 (41.2\%)}
& \textbf{84.6\%}
\\
OFT-Exit16
& 31/131 (23.7\%)
& 80.0\%
\\
\hline
\end{tabular}%
}
\end{table}

Policy switching, however, introduces a systems challenge. A synchronous
handoff exposes incoming-policy inference latency directly to robot control;
in our evaluation, the median exposed wait is 226.5~ms, equivalent to more than four nominal 50-ms control intervals at 20~Hz. Launching OFT asynchronously can hide this computation behind
remaining committed Adapter actions, but the robot continues moving during
inference, making the launch-time context stale at takeover. Heterogeneous
switching must therefore determine both \emph{when} to intervene and
\emph{how} to transfer control without blocking or context misalignment.

We present \textbf{AsyncMixVLA}, a runtime framework for proactive,
asynchronous switching between heterogeneous VLA policies. A learned
intervention trigger selectively escalates from VLA-Adapter to OpenVLA-OFT.
Once accepted, the remaining Adapter actions form a
\emph{committed-action bridge} that executes while OFT preparation proceeds
asynchronously. AsyncMixVLA then aligns the incoming policy's proprioceptive
and visual context to the anticipated takeover state using controller-calibrated
state roll-forward and an action-consistent visual-token residual. \textit{When the environment changes mid-execution, the runtime must decide whether another policy is more likely to complete the task from the current trajectory and, if so, prepare that policy for the state it will inherit.}

Unlike prior asynchronous VLA execution methods that primarily maintain
temporal consistency within a single deployed policy
~\cite{black2025rtc,tang2025vlash,wei2026f2fap}, AsyncMixVLA couples
selective transfer between independently trained policies with asynchronous,
context-aligned handoff. Our contributions are:

\begin{enumerate}
    \item \textbf{Complementarity-driven heterogeneous switching.}
    We characterize complementary failures of independently trained
    VLA policies and introduce a learned rescue-versus-harm trigger for
    selective escalation.

    \item \textbf{Proactive asynchronous handoff with context alignment.}
    We overlap incoming-policy inference with committed outgoing actions and
    align proprioceptive and visual context to the anticipated takeover state.

    \item \textbf{End-to-end robustness, responsiveness, and generalization.}
    Across 238 LIBERO-Long episodes, AsyncMixVLA improves perturbed success
    from \textbf{69.5\% to 77.1\%} over VLA-Adapter, achieves
    \textbf{84.9\% overall success} versus 84.5\% for synchronous switching,
    and reduces median handoff latency $L_{handoff}$ from \textbf{226.5~ms to 0.1~ms}.
    Without retraining or recalibration, it achieves \textbf{75.1\% success}
    across 289 episodes spanning three unseen perturbation families, showing generalization to unseen perturbations.
\end{enumerate}


\section{Related Work}
\label{sec:related_work}
\begin{figure*}[!ht]
    \centering
    \includegraphics[width=0.98\textwidth]{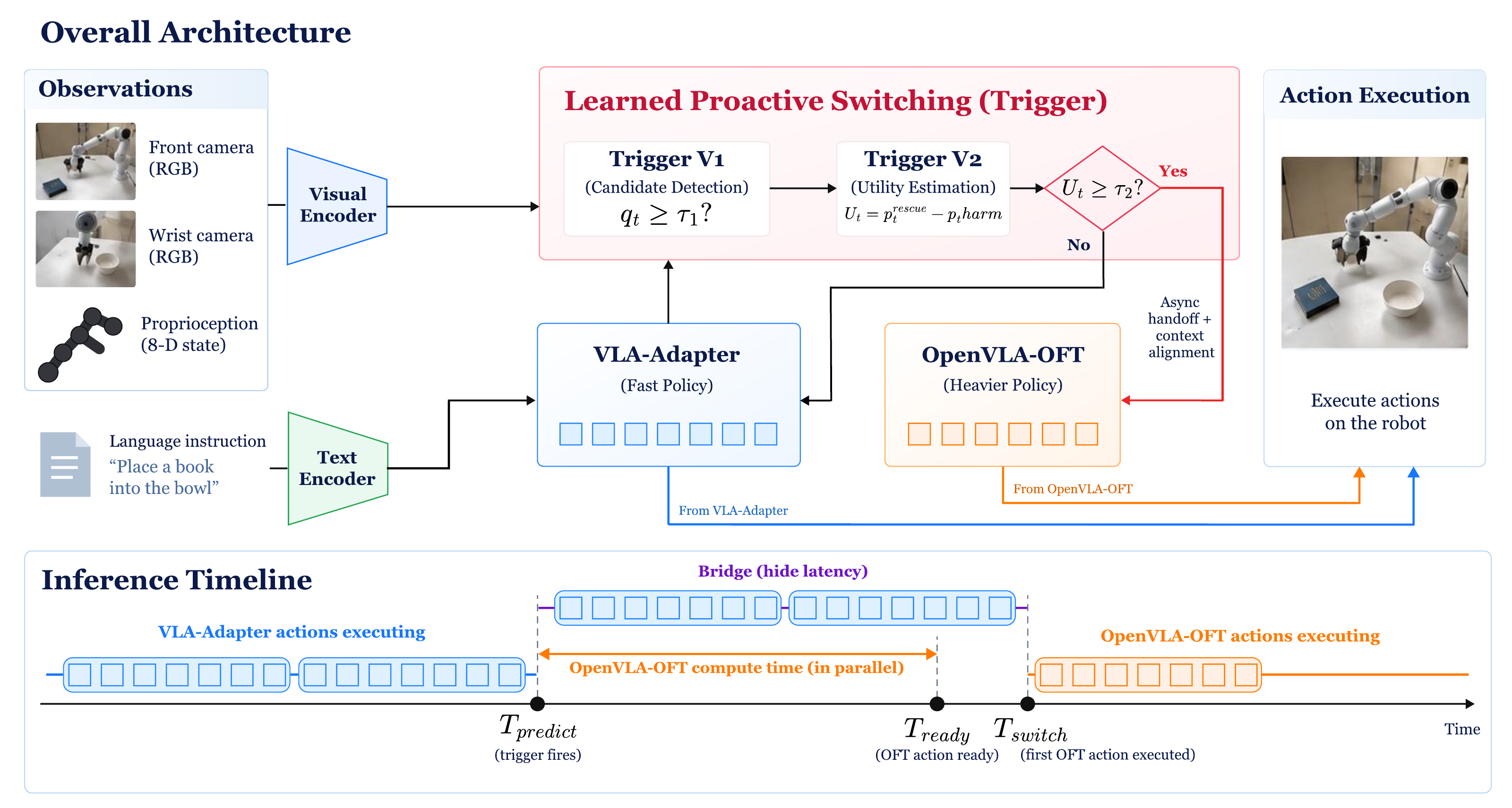}
    \caption{\textbf{Overview of AsyncMixVLA.}
    Trigger acceptance at $T_{\mathrm{predict}}$ starts OFT preparation while
    committed Adapter actions execute. OFT is conditioned on the predicted
    context at bridge completion, $T_{\mathrm{switch}}$. Its first action is
    issued at $T_{\mathrm{action}}$; fresh-observation replanning follows the
    initial prefetched takeover.}
    \label{fig:asyncmixvla_overview}
\end{figure*}

\noindent\textbf{VLA robustness and efficient inference.}
Robustness studies such as LIBERO-Plus~\cite{fei2025liberoplus} and
RobustVLA~\cite{guo2025robustvla} document VLA sensitivity to environmental
and observation perturbations. Efficient inference reduces policy cost
through compact backbones~\cite{wang2025vlaadapter,shukor2025smolvla},
layer/token pruning and feature
caching~\cite{yang2025efficientvla,jiang2025lightvla,xu2025vlacache}, parallel
action decoding~\cite{kim2025oft}, and adaptive early
exit~\cite{song2025ceedvla}. AsyncMixVLA complements these optimizations by
addressing when to invoke a higher-compute policy during execution. It
exploits complementary outcomes across independently trained policies while
retaining lightweight inference when escalation is unnecessary.

\noindent\textbf{Heterogeneous policy routing.}
RoboRouter~\cite{chen2026roborouter} selects policies for tasks using semantic
representations and accumulated execution experience. Pan
\emph{et al.}~\cite{pan2024switching} perform event-based transitions between
VLA and diffusion controllers for language-guided positioning and dexterous
grasping. These works establish precedents for heterogeneous selection and
switching during execution. AsyncMixVLA focuses on intervention under
disturbances: learned rescue-versus-harm utility decides whether to transfer
control from the current state, and asynchronous handoff reduces the transfer's exposed cost.

\noindent\textbf{Asynchronous policy execution.}
Action chunking~\cite{zhao2023act} amortizes inference but delays incorporation
of new observations; asynchronous execution~\cite{shukor2025smolvla}
overlaps prediction with ongoing control. Real-time chunking
(RTC)~\cite{black2025rtc} preserves committed actions through inpainting to
maintain chunk continuity. VLASH~\cite{tang2025vlash} conditions predictions
on rolled-forward robot states and incorporates temporal offsets during
fine-tuning. F2F-AP~\cite{wei2026f2fap} uses predicted object flow to anticipate
observations and a contrastive objective to align visual features. These
methods address temporal consistency for a deployed policy. AsyncMixVLA
shares their need to account for motion during inference, while addressing
selective handoff between independently trained controllers. Its incoming
policy is conditioned on the context expected after the outgoing policy's
committed actions. Visual alignment is trained through action consistency:
the frozen incoming policy's outputs under predicted and true handoff
contexts are matched, rather than directly matching visual features.
AsyncMixVLA thus combines intervention-utility routing, preparation during
committed outgoing actions, and alignment to the incoming controller's
anticipated takeover context.




\section{AsyncMixVLA}
\label{sec:method}

\subsection{System Formulation and Overview}
\label{sec:policy_formulation}

We consider long-horizon robotic manipulation with two independently trained
VLA policies: a lightweight default policy $\pi_A$ (VLA-Adapter) and a
higher-compute recovery policy $\pi_O$ (OpenVLA-OFT). The complementarity
characterized in Sec.~\ref{sec:introduction} motivates selective transfer
between these policies. Each episode begins under $\pi_A$ and permits at most
one transition to $\pi_O$, which then controls the remainder of the episode.

Given visual observations, measured proprioception, and a language instruction,
a query at step $r$ maps context $x_r=(I_r,s_r,\ell)$ to an $H$-action chunk,
\begin{equation}
    \mathbf{A}_r^i=\pi_i(x_r)
    =[a^i_{0|r},\ldots,a^i_{H-1|r}],
    \qquad i\in\{A,O\},
    \label{eq:action_chunk}
\end{equation}
where $H=8$ and $j$ indexes actions within the chunk.

Fig.~\ref{fig:asyncmixvla_overview} summarizes the runtime. A continuous
two-stage trigger selects intervention points. Once accepted, the unexecuted
suffix of the current Adapter chunk becomes a fixed
\emph{committed-action bridge}. The robot executes this bridge while an
asynchronous preparation path constructs an aligned takeover context and
queries OFT. Alignment accounts for the motion induced by the bridge, and
overlap reduces the incoming-policy latency exposed to robot control.
Takeover occurs after both bridge completion and response readiness.

\subsection{Continuous Learned Intervention Trigger}
\label{sec:trigger}

Episode-level complementarity does not determine whether switching helps
from the current state. AsyncMixVLA thus uses a two-stage V1--V2 cascade with only causal front/wrist RGB frames,
proprioception, Adapter actions and their history, and action-chunk position.
No privileged simulator state, object pose or identity, task ID, perturbation
metadata, future observation, or rollout outcome is available at runtime.

V1 combines temporal visual changes with robot state and the current Adapter
action. Its score $q_t\geq\tau_1$ proposes a candidate without committing a
switch. At each candidate, V2 combines causal rollout history, V1 visual
features, recent visual change, and $q_t$ to produce rescue and harm scores,
$p_t^{\mathrm{rescue}}$ and $p_t^{\mathrm{harm}}$. Their difference defines
intervention utility:
\begin{equation}
    U_t=p_t^{\mathrm{rescue}}-p_t^{\mathrm{harm}},
    \qquad \text{accept if }U_t\geq\tau_2,
    \label{eq:intervention_utility}
\end{equation}
Rejected candidates leave Adapter in control and re-arm V1 so the same fixed V2 model can reconsider later states. Acceptance records
$T_{\mathrm{predict}}$ and commits the one-way handoff; no further routing
decisions are made. Training labels, operating points, and the scope of these
scores are given in Sec.~\ref{sec:trigger_training}.

\subsection{Asynchronous Policy Handoff}
\label{sec:async_handoff}
\label{sec:oft_takeover}

Suppose acceptance occurs at step $t$, and $p$ is the next unexecuted position
in the Adapter chunk generated at step $r$. The committed bridge is
\begin{equation}
    \mathcal{B}_t=[a^A_{p|r},\ldots,a^A_{H-1|r}],
    \qquad |\mathcal{B}_t|=H-p.
    \label{eq:committed_bridge}
\end{equation}
Every action in this suffix executes. The bridge is fixed at
$T_{\mathrm{predict}}$, and no additional Adapter chunks are generated to
extend it. Concurrently, the preparation path constructs
$\hat{x}_{\mathrm{switch}}$ (Sec.~\ref{sec:handoff_alignment}) and obtains
the OFT response. 
An early response waits for bridge completion; a late
response exposes a residual wait before takeover. 
A longer committed bridge gives OFT preparation more time to overlap
with execution, but postpones the earliest possible OFT action even
if the response is ready. After a disturbance, the remaining Adapter
actions may no longer suit the changed scene. Takeover-context
alignment accounts for the motion induced by these actions, but does
not revise them.

Let $T_{\mathrm{ready}}$ denote response readiness,
$T_{\mathrm{switch}}$ completion of the final bridge action, and
$T_{\mathrm{action}}$ issuance of the first OFT action. The robot-visible
handoff latency is
\begin{equation}
    L_{\mathrm{handoff}}
    =T_{\mathrm{action}}-T_{\mathrm{switch}}.
    \label{eq:handoff_latency}
\end{equation}
We count a handoff as stall-free (next actions ready before bridge action completion) when
$T_{\mathrm{ready}}\leq T_{\mathrm{switch}}+T_{\mathrm{ctrl}}$, where
$T_{\mathrm{ctrl}}$ is one control period. Overlap hides context preparation and inference
behind useful Adapter execution without reducing OFT's intrinsic inference
time.

\textbf{Initial takeover and replanning.}
The initial prefetched OFT chunk contains eight actions. The deployed
configuration executes its first $k=4$ actions (200~ms of control steps),
discards the remaining four, and queries OFT from a fresh observation. This
limits open-loop execution from the predicted context while using multiple
prefetched actions. The truncation applies only to the initial takeover;
subsequent execution uses ordinary OFT replanning, with no return to Adapter.

\subsection{Handoff Context Alignment}
\label{sec:handoff_alignment}

\begin{figure}[!t]
    \centering
    \includegraphics[width=1.0\linewidth]{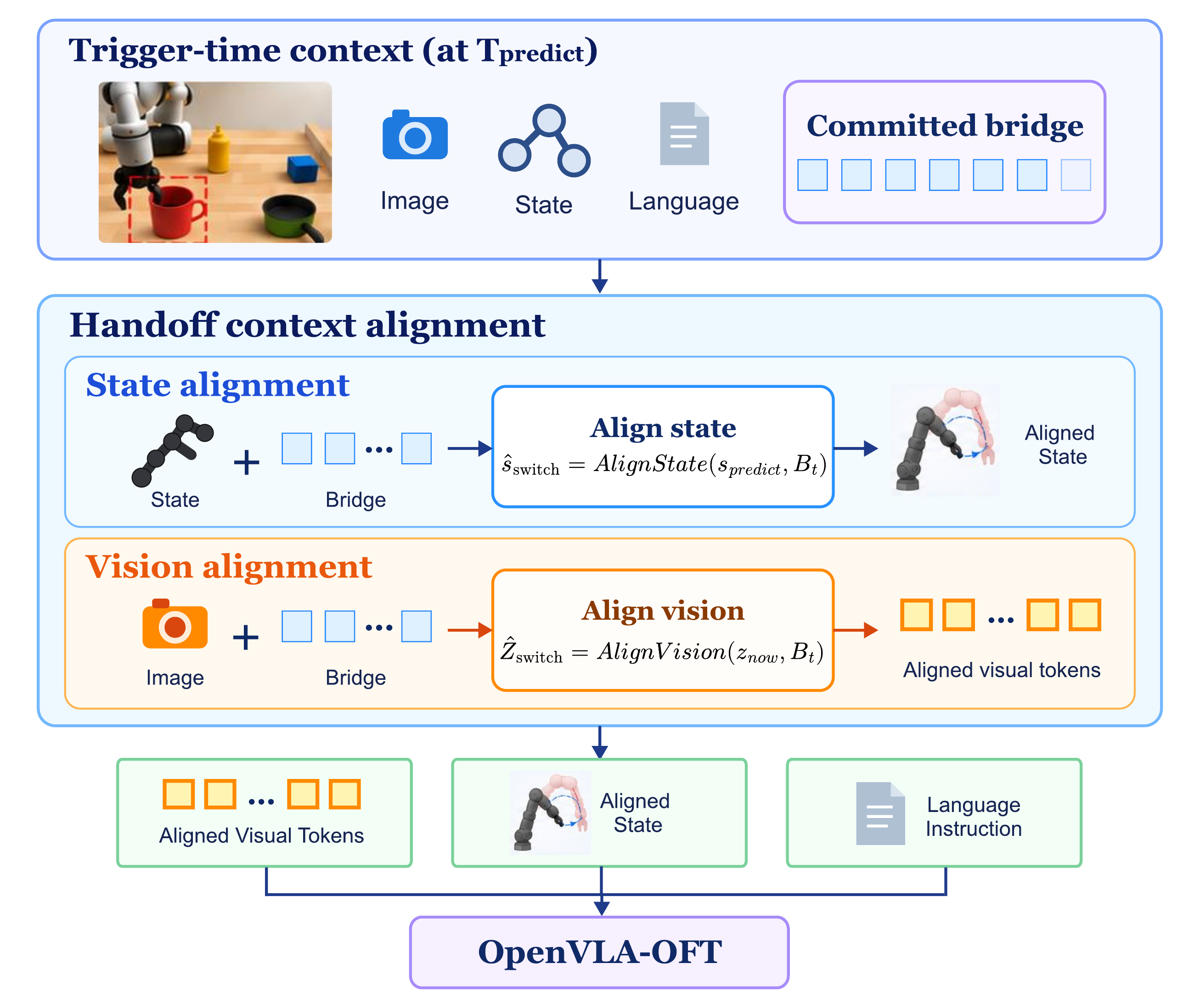}
    \caption{\textbf{Handoff context alignment.}
At trigger time, the current robot state and visual observation are aligned with the anticipated takeover state using the committed VLA-Adapter action bridge. State alignment predicts the robot state at handoff, while vision alignment projects the visual context into aligned visual tokens. The aligned state, visual tokens, and language instruction then form the takeover context for OpenVLA-OFT.}
    \label{fig:handoff_alignment}
\end{figure}

During asynchronous preparation, the robot moves through $\mathcal{B}_t$,
making the trigger-time context stale at bridge completion. Waiting for a
fresh observation would expose incoming inference latency. AsyncMixVLA
instead predicts $\hat{x}_{\mathrm{switch}}
=(\hat{Z}_{\mathrm{switch}},\hat{s}_{\mathrm{switch}},\ell)$ from the
trigger-time observation and known bridge
(Fig.~\ref{fig:handoff_alignment}). Here, $\hat{Z}_{\mathrm{switch}}$
represents projected visual tokens supplied through OFT's existing token
interface, rather than a synthesized image.

\subsubsection{Controller-Calibrated Proprioceptive Alignment}
The measured state $s_{\mathrm{predict}}\in\mathbb{R}^{8}$ contains
end-effector position, axis-angle orientation, and two gripper values.
We first compute $\tilde{s}_{\mathrm{switch}}
=\mathcal{F}(s_{\mathrm{predict}},\mathcal{B}_t)$ using LIBERO controller
semantics: accumulate translation, compose rotations, and propagate gripper
commands. Because the OSC-Pose impedance controller does not fully realize
each commanded displacement within one step, calibrated component-wise gains
correct the roll-forward estimate:
\begin{equation}
    \hat{s}_{\mathrm{switch}}=s_{\mathrm{predict}}
    +\mathbf{g}\odot
    (\tilde{s}_{\mathrm{switch}}-s_{\mathrm{predict}}).
    \label{eq:proprioceptive_alignment}
\end{equation}

\subsubsection{Action-Consistency Visual Alignment}
Let $Z_{\mathrm{now}}=E_{\mathrm{OFT}}(I_{\mathrm{predict}})$ denote the
current projected visual tokens. A lightweight predictor $f_\phi$, conditioned
on pooled OFT visual features and a summary of the committed bridge, produces
a residual shared across all tokens:
\begin{equation}
\Delta z
=
f_{\phi}\!\left(
\operatorname{MeanPool}(Z_{\mathrm{now}}),\, b_t
\right),
\qquad
\hat{Z}_{\mathrm{switch}}
=
Z_{\mathrm{now}}
+
\mathbf{1}\Delta z^{\top}.
\label{eq:visual_alignment}
\end{equation}

The residual is trained offline with OFT frozen using an
\emph{action-consistency} objective. Writing $\pi_O(Z,s,\ell)$ for OFT
evaluated through its projected-token interface, the loss matches the policy
output under the predicted context to its output under the true post-bridge
context:
\begin{equation}
    \mathcal{L}_{\mathrm{AC}}=
    \left\|
    \pi_O(\hat{Z}_{\mathrm{switch}},\hat{s}_{\mathrm{switch}},\ell)
    -\pi_O(Z_{\mathrm{true}},s_{\mathrm{true}},\ell)
    \right\|_2^2.
    \label{eq:action_consistency}
\end{equation}
Thus, supervision targets OFT behavior rather than latent reconstruction.
The true post-bridge tokens and state are used only during training. At
deployment, visual alignment depends solely on $Z_{\mathrm{now}}$ and
$\mathcal{B}_t$, while proprioceptive alignment also uses the measured
state. OFT's backbone, language conditioning, proprioceptive pathway, and
action head remain frozen.


\section{Experimental Setup}
\label{sec:experimental_setup}

\noindent\textbf{Environment and perturbations.}
\label{sec:environment}
\label{sec:perturbation_protocol}
We evaluate on the ten long-horizon tasks of
\textbf{LIBERO-Long}~\cite{liu2023libero} at 20~Hz
($T_{\mathrm{ctrl}}=50$~ms), with at most $T_{\max}=520$ environment actions,
7-D actions, and the native $H=8$ action horizon. 
The 20-Hz rate denotes the simulation control timestep rather than enforced wall-clock pacing: simulation state advances only through executed \texttt{env.step()} actions. It therefore continues to evolve during committed bridge actions but does not evolve during residual wall-clock waiting after bridge completion. We report this residual as exposed handoff wait while evaluating task recovery under action-synchronous dynamic perturbations.
Frozen VLA-Adapter and
OpenVLA-OFT checkpoints serve as default and recovery policies; only the
trigger and alignment components are trained or calibrated. 

\textbf{Perturbation Benchmark design (238 cases):} The primary
benchmark contains 120 clean episodes (12 fixed initial states $\times$ 10 tasks) and
118 feasible matched perturbed counterparts (two excluded by feasibility
constraints), giving 238 total. The failure-overlap characterization in
Sec.~\ref{sec:introduction} uses a separate set of 500 matched perturbed
rollouts, with no clean episodes, solely to measure episode-level policy
complementarity; it is distinct from this final 238-episode evaluation.
Perturbations displace a task-relevant movable source object (e.g., juice can) by 5~cm in-plane over
five control steps. A frozen manifest fixes object, onset, direction,
magnitude, and duration across methods. The perturbation hook precedes
every environment action, including bridge and post-takeover actions, so
overlapping disturbances continue through handoff. All methods share seeds,
checkpoints, action processing, manifests, and horizons.

\noindent\textbf{Training and implementation.}
\label{sec:trigger_training}
\label{sec:alignment_configuration}
The trigger is trained offline and frozen before evaluation. V1 uses
$64\times64$ front/wrist RGB images with temporal differences at lags
$\{1,4,16\}$; calibration on trials 32--39 selects $\tau_1=0.92296$ with
persistence one. V2 uses 199 fit examples from trials 12--31, calibration
trials 32--39, and development trials 40--49. Matched continuations from the
same causal prefix either retain Adapter or invoke the then-frozen
AsyncMixVLA handoff; binary outcomes define $\Delta y=y_M-y_A$, with
$+1$, $-1$, and $0$ denoting rescue, harm, and tie. Ties are retained;
regularized logistic rescue/harm heads use no class balancing, with features
standardized using fixed training statistics and clipped to $[-10,10]$.
Calibration selects $\lambda=0.1$ and $\tau_2=-0.1$ to maximize paired
success while retaining at least 90\% of direct-V1 rescues without increasing
harms; ties favor fewer switches, then stronger regularization. 
We fit V2 on first intervention candidates using outcomes from paired continuations that either retain Adapter or invoke the
AsyncMixVLA handoff. In Section\ref{sec:learned_trigger_results}, we deploy the frozen V2 with
continuous candidate re-arming and execute the first $k=4$ actions
of OFT's initial chunk at takeover. V2 outputs serve as
intervention-utility scores.
Proprioceptive alignment uses gains $g_{\mathrm{pos}}=0.252$,
$g_{\mathrm{ori}}=0.240$, and $g_{\mathrm{gripper}}=0$.
Visual alignment concatenates mean-pooled 4096-D OFT tokens with a 22-D
bridge descriptor (mean/first/last actions and normalized length) and uses a
two-layer MLP of hidden width 64 to predict a shared 4096-D residual, trained
with Eq.~\ref{eq:action_consistency} while OFT remains frozen.
The visual-alignment predictor is fitted on 1,668 deploy-matched bridge
samples from trials 12--39 across all ten LIBERO-Long tasks. An additional
597 samples from trials 40--49 are reserved for development, while the final
238-episode evaluation uses disjoint trials 0--11.

\noindent\textbf{Baselines.}
\label{sec:compared_methods}
We compare Adapter-only and OFT-only execution with synchronous switching,
naive asynchronous switching, and AsyncMixVLA. All switching methods share
identical frozen trigger decisions, committed bridges, and $k=4$ takeover
horizons, without generating additional Adapter chunks to extend the bridge.
Synchronous switching queries fresh context after bridge completion; naive
asynchronous switching starts preparation at $T_{\mathrm{predict}}$ using
trigger-time context; AsyncMixVLA starts at the same boundary and constructs
the aligned takeover context before querying OFT.


\noindent\textbf{Metrics and timing.}
\label{sec:metrics}
\label{sec:runtime_timing}
We report overall, clean, and perturbed task success rates; Adapter failures
rescued and successes retained; and the fraction of episodes invoking OFT.
OFT invocation rate measures intervention selectivity, not compute or energy
savings, since OFT may execute for the remainder of an episode after takeover.
Responsiveness uses median/$p95$ handoff latency
(Eq.~\ref{eq:handoff_latency}) and the no-stall criterion of Sec.~\ref{sec:async_handoff}. All timing experiments run on a single NVIDIA RTX 3090 GPU, with VLA-Adapter and OpenVLA-OFT using BF16 inference as separate single-worker services within the same GPU allocation. Thus, asynchronous measurements include GPU contention between the concurrently resident policies. Under the final learned trigger, latency statistics are computed over the same 109 handoffs for each switching method.
Handoff latency is measured in wall-clock time, independently
of the simulator's action-synchronous progression. Monotonic
timing uses \texttt{time.perf\_counter()} inside the end-to-end
runtime: $T_{\rm predict}$ precedes all alignment and OFT
preparation, $T_{\rm ready}$ follows complete response receipt
and post-processing, and $T_{\rm switch}$ marks completion of
the final bridge \texttt{env.step()}.


\section{Experimental Results}
\label{sec:results}

\subsection{End-to-End Learned Policy Switching}
\label{sec:learned_trigger_results}


\begin{table*}[!t]
\centering
\caption{
End-to-end performance of static and heterogeneous policy-switching strategies.
\textit{Rescued Adapter Failures} reports the number of Adapter failures
recovered by each method, while \textit{Retain Rate} measures the fraction of
Adapter-successful episodes that remain successful. Switch/OFT Invocation
reports the fraction of episodes in which OFT is executed.
}
\label{tab:end_to_end_results}
\begin{tabular*}{\textwidth}{@{\extracolsep{\fill}}lcccccc@{}}
\hline
\textbf{Method} &
\begin{tabular}[c]{@{}c@{}}\textbf{Overall}\\\textbf{SR}\end{tabular} &
\begin{tabular}[c]{@{}c@{}}\textbf{Clean}\\\textbf{SR}\end{tabular} &
\begin{tabular}[c]{@{}c@{}}\textbf{Perturbed}\\\textbf{SR}\end{tabular} &
\begin{tabular}[c]{@{}c@{}}\textbf{Rescued Adapter}\\\textbf{Failures}\end{tabular} &
\begin{tabular}[c]{@{}c@{}}\textbf{Retain}\\\textbf{Rate}\end{tabular} &
\begin{tabular}[c]{@{}c@{}}\textbf{Switch/OFT}\\\textbf{Invocation}\end{tabular} \\
\hline

VLA-Adapter only
& 81.1\%
& 92.5\%
& 69.5\%
& 0/45 (0.0\%)
& \textbf{193/193 (100.0\%)}
& 0.0\% \\

OpenVLA-OFT only
& 83.2\%
& 92.5\%
& 73.7\%
& \textbf{26/45 (57.8\%)}
& 172/193 (89.1\%)
& 100.0\% \\

Synchronous switching
& 84.5\%
& 92.5\%
& 76.3\%
& 16/45 (35.6\%)
& 185/193 (95.9\%)
& 45.8\% (109) \\

Naive asynchronous
& 83.2\%
& 92.5\%
& 73.7\%
& 16/45 (35.6\%)
& 182/193 (94.3\%)
& 45.8\% (109) \\

\textbf{AsyncMixVLA (Ours)}
& \textbf{84.9\%}
& 92.5\%
& \textbf{77.1\%}
& 17/45 (37.8\%)
& 185/193 (95.9\%)
& 45.8\% (109) \\

\hline
\end{tabular*}
\end{table*}

\begin{figure}[!t]
\centering
\includegraphics[width=\columnwidth]{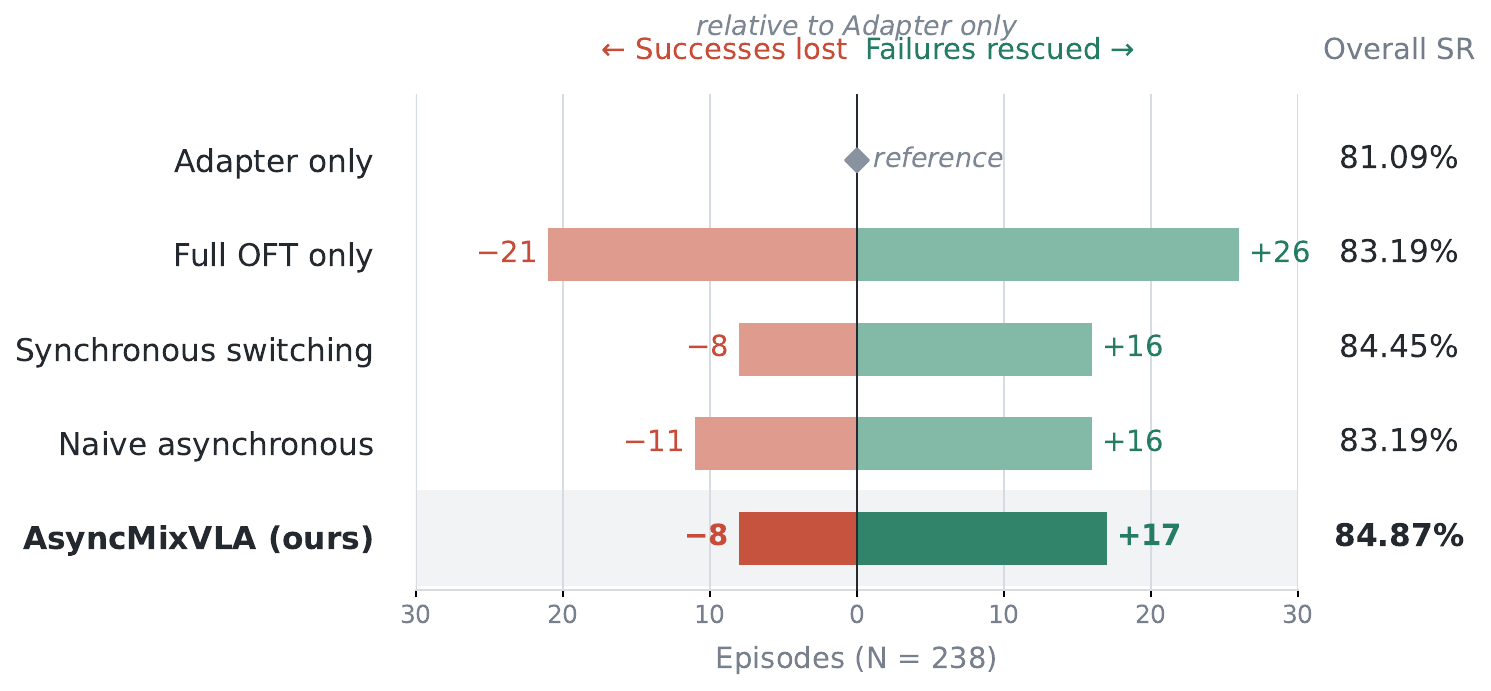}
    \caption{
        \textbf{Rescue--retention trade-off relative to VLA-Adapter.}
        Rightward red bars denote rescued Adapter failures; leftward green bars denote Adapter successes converted to failures.
    }
\label{fig:rescue_retention_tradeoff}
\end{figure}




We evaluate the learned trigger with frozen continuous V1--V2 cascade end-to-end on the same perturbation benchmark (Table~\ref{tab:end_to_end_results}).
With the learned trigger, AsyncMixVLA completes 91 of 118 perturbed
episodes (77.1\%), compared with 82 (69.5\%) for VLA-Adapter alone.
All policies complete 111 of 120 clean episodes successfully (92.5\%). Thus, AsyncMixVLA gains nine completions overall while invoking OpenVLA-OFT in only 109 of
238 episodes (45.8\%). Fig.~\ref{fig:rescue_retention_tradeoff} shows the rescue--retention trade-off
behind this result. Compared with always-on OpenVLA-OFT, AsyncMixVLA
rescues nine fewer recorded VLA-Adapter failures (17 versus 26) but
retains 13 more recorded successes (185 versus 172), completing four
more episodes overall. Selective intervention therefore
balances recovery opportunities against intervention harm, emphasizing the aforementioned \emph{rescue--retention
trade-off}.

At the same intervention points, AsyncMixVLA and naive asynchronous
handoff differ in the context supplied to the first OpenVLA-OFT
query. Five episodes succeed only with takeover-aligned context and
one succeeds only with launch-time context. We therefore interpret
AsyncMixVLA as offering comparable task success to synchronous
switching, rather than statistically superior task success. Its primary systems
advantage is instead the cost of performing the handoff, which we examine next.


\subsection{Responsiveness and Handoff Latency}
\label{sec:latency_results}

\begin{table*}[!t]
    \centering
    \caption{Task success and exposed wait under identical handoff decisions.
    Perturbed success is measured over 118 episodes, overall success over
    238 episodes, and exposed wait over 109 handoffs per method.}
    \label{tab:handoff_success_wait}
    \small
    \renewcommand{\arraystretch}{1.12}
    \begin{tabular}{@{}llcccc@{}}
        \hline
        & & \multicolumn{2}{c}{\textbf{Task success (\%)}} &
            \multicolumn{2}{c}{\textbf{Exposed handoff latency (ms)}} \\
        \cline{3-4}\cline{5-6}
        \textbf{Handoff method} & \textbf{Context for first OFT query} &
        \textbf{Perturbed} & \textbf{Overall} &
        \textbf{Median} & \textbf{p95} \\
        \hline
        Synchronous        & Fresh, after bridge  & 76.3 & 84.5 & 226.5 & 363.8 \\
        Naive asynchronous & Observed at launch   & 73.7 & 83.2 &   0.1 & 200.1 \\
        AsyncMixVLA        & Aligned to takeover  & 77.1 & 84.9 &   0.1 & 224.5 \\
        \hline
    \end{tabular}
\end{table*}

Table~\ref{tab:handoff_success_wait} shows that asynchronous handoff reduces median exposed handoff latency  from 226.5\,ms to 0.1\,ms. AsyncMixVLA has slightly higher
tail latency than naive asynchronous switching due to alignment overhead, but
improves perturbed success and retention with identical learned-trigger
decisions (Table~\ref{tab:end_to_end_results}).


\subsection{Alignment Ablation and Action Fidelity}
Table~\ref{tab:action_fidelity} isolates the two alignment components we add. For each of the 109 handoffs, we compare the first OFT chunk produced by each asynchronous method with the chunk produced by synchronous switching, which receives the fresh observation. Visual alignment accounts for most of the reduction in executed-action deviation, and combining it with state alignment gives the lowest deviation.

\begin{table}[t]
\centering
\caption{L2 deviation of OFT's four executed takeover actions from the synchronous reference over 109 handoffs (excluding the gripper action; lower is better).}
\label{tab:action_fidelity}
\footnotesize
\begin{tabular}{lc}
\hline
\textbf{Context for first OFT query} & \textbf{Executed-action L2} \\
\hline
Naive asynchronous    & 0.1760 \\
State alignment only  & 0.1750 \\
Vision alignment only & 0.1687 \\
AsyncMixVLA (both)    & \textbf{0.1676} \\
\hline
\end{tabular}
\end{table}

\subsection{Zero-Shot Generalization to Unseen Perturbations}
\label{sec:zero_shot_perturbations}

Finally, we test whether the learned selective-switching behavior transfers
beyond the translation disturbance used for development. Without retraining or
recalibration, we evaluate the frozen system on three unseen perturbation
families: object rotation up to $\pm28.6^\circ$, 5-cm target displacement (e.g., moving the basket, not the source object), and
3-cm post-grasp slip.

\begin{table}[!t]
\centering
\caption{
Zero-shot evaluation on perturbation families not used in the perturbation benchmark. Values denote task success rate.
}
\label{tab:zero_shot_perturbations}
\setlength{\tabcolsep}{4.2pt}
\renewcommand{\arraystretch}{1.12}
\resizebox{\columnwidth}{!}{%
\begin{tabular}{lcccc}
\hline
\textbf{Perturbation}
&
\textbf{Episodes}
&
\textbf{VLA-Adapter}
&
\textbf{OpenVLA-OFT}
&
\textbf{AsyncMixVLA}
\\
\hline

Object rotation, $\pm 28.6^\circ$
& 118
& 83.9\%
& \textbf{85.6\%}
& 84.8\%
\\

Target displacement, 5~cm
& 60
& 70.0\%
& 58.3\%
& \textbf{73.3\%}
\\

Post-grasp slip, 3~cm
& 111
& 58.6\%
& 57.7\%
& \textbf{65.8\%}
\\
\hline

\textbf{All unseen perturbations}
& \textbf{289}
& 71.3\%
& 69.2\%
& \textbf{75.1\%}
\\
\hline
\end{tabular}%
}
\end{table}

Table~\ref{tab:zero_shot_perturbations} shows that policy complementarity
persists under these unseen conditions: neither constituent policy dominates
across all three perturbation families. OpenVLA-OFT is strongest on object
rotation (85.6\%), whereas VLA-Adapter substantially exceeds it under target
displacement (70.0\% vs.\ 58.3\%). AsyncMixVLA achieves the highest observed
SR on target displacement (73.3\%) and post-grasp slip (65.8\%), while
remaining close to OpenVLA-OFT on rotation (84.8\% vs.\ 85.6\%).
Across all 289 unseen-perturbation episodes, AsyncMixVLA reaches 75.1\%
success, compared with 71.3\% for VLA-Adapter and 69.2\% for
OpenVLA-OFT. These results suggest that the selective heterogeneous cascade transfers beyond
the translation disturbance used for development without retraining or
recalibration.








\section{Conclusion}
We presented AsyncMixVLA, a runtime framework for selective asynchronous
switching between heterogeneous VLA policies. The approach is motivated by
complementary failure behavior between VLA-Adapter and OpenVLA-OFT, uses a
continuous learned rescue-versus-harm trigger to decide when escalation is
worthwhile, overlaps OpenVLA-OFT preparation with already committed
VLA-Adapter actions, and aligns the incoming policy's proprioceptive and
visual context to the anticipated handoff state.

On the frozen learned-trigger evaluation, AsyncMixVLA reaches 84.9\% overall
and 77.1\% perturbed success while retaining 185/193 VLA-Adapter successes.
Its median exposed handoff wait is 0.1~ms, compared with 226.5~ms for
synchronous switching, while zero-shot experiments show improved aggregate
success across three unseen perturbation families. These results show that proactive asynchronous handoff can substantially reduce typical exposed switching latency while preserving the robustness benefit of selective heterogeneous policy execution.


\bibliographystyle{IEEEtran}
\bibliography{references}

\end{document}